\documentclass[conference]{IEEEtran}
\IEEEoverridecommandlockouts
\usepackage{cite}
\usepackage{amsmath,amssymb,amsfonts}
\usepackage{algorithmic}
\usepackage{graphicx}
\usepackage{textcomp}
\usepackage{xcolor}
\usepackage{adjustbox}
\usepackage{soul}
\usepackage[font=small]{caption}
\def\BibTeX{{\rm B\kern-.05em{\sc i\kern-.025em b}\kern-.08em
    T\kern-.1667em\lower.7ex\hbox{E}\kern-.125emX}}

\begin{document}

%\font\myfont=cmr12 at 14pt
% a specific functionality ....
%\title{\myfont \textbf{Automated Drug-Related Information Extraction from French Clinical Documents: ReLyfe  Approach}\\

\title{Automated Drug-Related Information Extraction from French Clinical Documents: ReLyfe  Approach}

\makeatletter
\newcommand{\linebreakand}{%
  \end{@IEEEauthorhalign}
  \hfill\mbox{}\par
  \mbox{}\hfill\begin{@IEEEauthorhalign}
}
\makeatother

\author{
\IEEEauthorblockN{Azzam Alwan}
\IEEEauthorblockA{\textit{R\&D Department} \\
\textit{ReLyfe - Medical Intelligence }\\
Paris, France\\
\small{email: azzam.alwan@relyfe.com}}
\and
\IEEEauthorblockN{Maayane Attias}
\IEEEauthorblockA{\textit{Computer Science} \\
\textit{Ecole Polytechnique}\\
Palaiseau, France \\
\small{email: maayane-lea.attias@polytechnique.edu}
}
\linebreakand
\IEEEauthorblockN{Larry Rubin}
\IEEEauthorblockA{\textit{R\&D Department} \\
\textit{BeCareLink}\\
New York, United States \\
\small{email: larry.rubin@becarelink.com}}

%\IEEEauthorblockN{Marc Salomon, MD.}
%\IEEEauthorblockA{\textit{R\&D Department} \\
%\textit{ReLyfe - Medical Intelligence}\\
%Paris, France \\
%\small{email: marc.salomon@relyfe.com}}
%\and
\and
\IEEEauthorblockN{Adnan El Bakri}
\IEEEauthorblockA{\textit{R\&D Department} \\
\textit{ReLyfe - Medical Intelligence}\\
Paris, France \\
\small{email: ceo@relyfe.com}}
}
\maketitle

\begin{abstract}
Structuring medical data in France remains a challenge mainly because of the lack of medical data due to privacy concerns and the lack of methods and approaches on processing the French language.
One of these challenges is structuring drug-related information in French clinical documents.
To our knowledge, over the last decade, there are less than five relevant papers that study French prescriptions. This paper proposes a new approach for extracting drug-related information from French clinical scanned documents while preserving patients' privacy. In addition, we deployed our method in a health data management platform where it is used to structure drug medical data and help patients organize their drug schedules. It can be implemented on any web or mobile platform. This work closes the gap between theoretical and practical work by creating an application adapted to real production problems. It is a combination of a rule-based phase and a Deep Learning approach. Finally, numerical results show the outperformance and relevance of the proposed methodology.

%In this paper we propose a new web/mobile functionality to extract drug related information form french prescription scanned document.  The objective is to extract 
%In this paper, we propose one of our several AI solution that we have developed in the research department of our startUps "Relyfe AI".  Our main objective is to develop a web/mobile application to extract drugs medical information from clinical texts written in French and then store them in databases centralized around the patient where he can be access to it wherever he is via his smartphone.
\end{abstract}


\section{Introduction}
Today, the main source of mistakes in medicine is due to the lack of information that doctors have on their patients, especially when multiple doctors treat the same patient. The information flow has to be as efficient and smooth as possible, especially when patient is transferred from one doctor to another. When making a medical decision, it is necessary to avoid wrong actions that are incoherent with the patient’s status, such as incompatibility between medication and their medical condition. Consequently, doctors need to have access to all of the patient’s information and background. However, this is not always the case since healthcare systems do not have patient’s information gathered all in one place: medical information only exists in silos. For example, the main source of medication errors is related to wrong drug prescriptions having severe consequences on a patient’s health \cite{tariq2020medication}.

Based on the arguments mentioned so far, we see that there is a crucial need to gather patients’ information in one unique system/platform to be easily retrieved, understood, and shared. Electronic Health Records (EHRs) provide healthcare workers with a better understanding of the patient thanks to the stored information. However,  the information not only needs to be stored, it also needs to be structured to be used efficiently. Indeed, when clinical data is extracted from a document and structured, it is easier to be examined and interpreted. Today, 80\%  of relevant clinical information exists only in an unstructured form \cite{jouffroy2021hybrid}, which results in a massive loss of helpful information. We decided to focus in this paper on the structuration of medical prescriptions. They contain valuable information about a patient’s drug history and, when structured, can be used for pharmaco-vigilance and epidemiology. Manual extraction would be too difficult and time-consuming, which explains why clinical Natural Language Processing (NLP) and entity extraction are of genuine interest in the field of research. Considering the lack of solutions available in the French health care system, we propose our algorithm and put it at the service of any other solution that enhances the health care system.

The rest of the paper is organized as follows. The problem formulation is presented in Section \ref{Relatedwork}. In Section \ref{MaterialsMethods}, the proposed methodology is described. Experimental results evaluating the efficiency of the proposed method are presented in Section \ref{EvaluationResult}. Finally, conclusions are drawn in Section \ref{conclusion}.

%Considering the lack of solutions available in the French health care system, we decided to create our own algorithm. In our work, we developed a hybrid approach to extract drug-related information from French prescriptions. We contributed to research in this field by training a classifier to decide in which category a sentence lies (drug, posology or useless sentence) and then applying a rule based approach crafted accordingly to the identified category. The sentence classification done prior to the rule-based approach prevents from assigning wrong entities to the text. It contributes to the great performance achieved (put a score?) when distinguishing the frequency of the prescription with the strength. Most of the other models usually lack sensitivity on this point and confuse those two entities. In addition, we then link the sentences classified as drugs to a unique ID in the Vidal database. That way, the drug is recognized in an international database and additional information are fetched. By matching the extracted drug with the Vidal database using a matching algorithm, we were also able to deal with the potential errors occurring when running the OCR on the scanned document. Indeed, even if the drug name is not complete, since we use a "closest-matching" algorithm we are still able to recover the official drug name from the Vidal database query. Finally, the geometric approach we used to match a drug to its associated posology also contibutes to efficiency of our model.

\section{Related Work}\label{Relatedwork}

The majority of NLP research work on medical data has been carried out on texts/documents in English, whether it is to structure or analyze them \cite{gold2008extracting,  shen2020natural,abacha2011automatic,lupde2018extracting, patrick2010high, neveol2018clinical, sheikhalishahi2019natural }. In fact, the need to structure medical documents in any language is as strong as in English.  In French specifically, many policies advocate for a better healthcare system while there is still a lack of structured data and research. One of the principal methodologies in NLP applied to medical data is the Named Entity Recognition (NER), whose purpose is to extract relevant medical information from unstructured medical text. To our knowledge, since 2010, there have been less than five pertinent papers talking about the extraction of drug-related information from French clinical prescriptions \cite{deleger2010extracting, lerner2020terminologies, jouffroy2021hybrid}. In addition, none of them is open-source, or a paid service or can be used as a real application. One of the most known method applied in practice is the Amazon Web services (AWS) medical service (Amazon Comprehend Medical); however, its initial release can only detect medical entities in English texts \cite{bai2021clinical}. Therefore, the idea is to use our expertise to achieve something similar for French medical documents. The major approaches in NER for medication are based on lexicon/rules and machine learning and can be combined into a hybrid model.

First, lexicon-based approaches use predefined lexicons or regular expressions to match parts of the text to recognize predefined entities. They aim to model expert knowledge with dictionaries. Examples of such approaches for English clinical texts are MedEx \cite{xu2010medex} and MedXN \cite{sohn2014medxn}.  These models are built to recognize seven categories: drug name, route, frequency, dosage, strength, form, duration. They have shown a good performance compared to other methods. Indeed, MedXN gives a F1-score of 0.975 for medication name and over 0.90 for attributes for a dataset of 397 medication mentions. Regarding the rule-based approaches for French clinical texts, the extraction also relies on a system with specialized lexicons and extraction rules in a similar approach to English. Some research has shown that the same methods in English can be applied to French \cite{deleger2010extracting}, however, they does not perform as well as in  English.

Regardless of the considered language, in a rule-based approach, the rules have to be extremely fine-tuned to fit the entities that will be extracted. Most researchers agree that this can be challenging when working with a very large dataset\cite{kreimeyer2017natural}. Additionally, it gives poor results when applied to texts deviating from the ones referenced in the dictionary. However, in case the whole entity is well predefined, this method provides precise results. That is why, in our approach, we have applied this method to get the drug name: we are connected to a database with all drug names in the French market \cite{medicaments}. In addition, based on the solid expertise of our physicians, we have been able to design and build rules and patterns covering drug-related information with a variety of abbreviations, grammatical errors and physicians' common mistakes.

There exists a second approach that uses machine learning in order to extract entities (dosage, frequency, duration, route, drug name) from clinical texts in \cite{alfattni2021extracting}. The system proposed in \cite{alfattni2021extracting} is a NER model relying on a bidirectional long-short term memory with conditional random fields (BiLSTM-CRF) architecture composed of 3 different layers: embedding layer, bidirectional long-short term memory layer, and conditional random fields layer. Different deep learning-based approaches were explored. All of them rely on a bidirectional long-short term memory with conditional random fields but with various embeddings such as word embedding, character embedding, and semantic-feature embedding. This system achieved encouraging results and demonstrated the feasibility of using deep learning methods to extract medication information from raw clinical texts. Several other research papers have proposed BiLSTM with conditional random fields for their NER model \cite{lample2016neural, sadikin2016new}. However, this type of approach requires working with words and phrases that have meaning in their sequence. However, in French prescriptions, the information is presented as a distinct set of words, where each set is generally made up of three or four words that are always in the same order. In order to leverage the advantages of these methods and the rule-based approaches, researchers combined both methods, which shows a better performance. An example of such a combination can be found in \cite{patrick2010high}, where the authors use a conditional random field for the NER model along with a support vector machine extracting related entities combined with a rule-based context engine. These approaches were designed for text written in English. To the best of our knowledge, there are only a few research works done for French clinical texts. In \cite{jouffroy2021hybrid}, a hybrid system combining a rule-based approach with contextual word embedding trained on clinical data with a deep recurrent neural network was developed, and it outperforms other approaches based on a token-level evaluation.
%In this work, we have used a combination of deep learning technique and a rule-based approach to do the NER, but differently. A deep learning approach was used to classify the described medical sentences. A rule-based approach has been implemented to extract the required entities. Indeed, the text classification model has indirectly improved the output of our NER model because it simply leads us to apply the rule-based models to the appropriate sentence containing the required information. That means reducing the model error that comes from applying rules-based patterns to a confusing sentence that may or may not contain what we are looking for. In the following section, a detailed explanation will be presented for each part of our methods.

In this work, we combine deep learning techniques and rule-based approaches to create our NER model from a different perspective. A deep learning approach is used to classify sentences into three categories (drug, posology, or useless sentence), and then a rule-based approach is crafted accordingly to the identified category. Indeed, applying the sentence classification model beforehand improves the output of our global model since it applies the rule-based extractor on the appropriate sentence containing the required information, i.e., prevents applying rules-based patterns on a confusing sentence that may or may not contain the desired entity. In addition, we then link the sentences that are classified as drugs to a unique ID in the Vidal database\cite{vidal}. That way, the drug is recognized in an international database, and additional information is fetched. By matching the extracted drug with the Vidal database, we can also deal with the potential errors occurring when running the Optical Character Recognition (OCR) on the scanned document. Indeed, even if the drug name is not complete, our algorithm can still recover the official drug name. Finally, we designed a geometric approach to build a relation extractor system that matches a drug and its associated posology. In the following section, we present a detailed explanation of each part of our method.

\section{Materials and Methods}\label{MaterialsMethods}
This section presents in detail the explicit steps and tools used and applied to achieve our desired goal. First, we describe the employed data. Then, we define the tools and libraries used to annotate the data and establish its utility. Also, we determine the referenced databases used in the project. Lastly, we explain the steps developed in our algorithms to extract a drug and its related information (dose, frequency, duration, and comment) and how we connect them.

\begin{figure*}[htb]
\centerline{\includegraphics[scale=0.25]{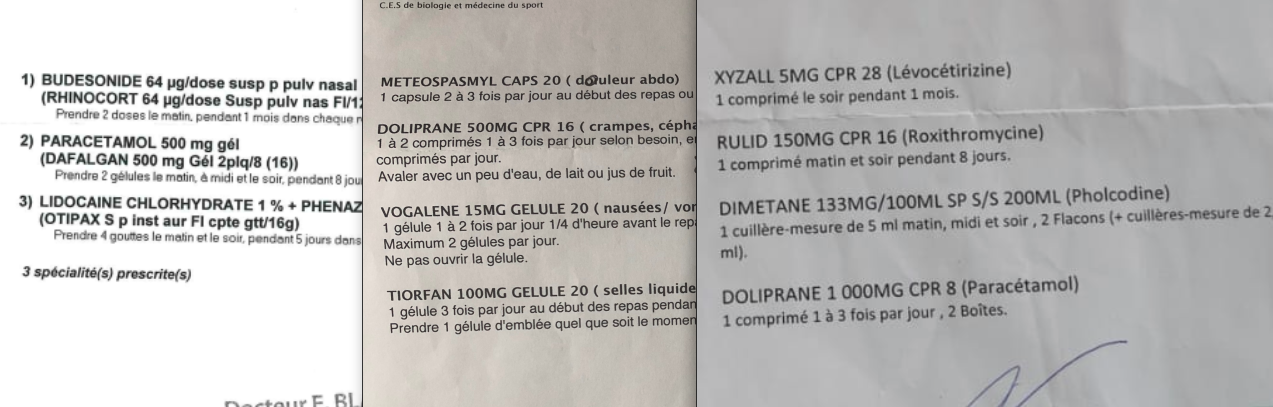}}
\caption{Example of scanned documents. These prescriptions were scanned with a smartphone camera. They vary in terms of quality, orientation, and color. Some documents contain the drug and its equivalent in case it is not available. Our algorithm should be clever enough to detect this case and choose only one of them.}
\label{scanned_documents}
\end{figure*}

\subsection{Data}\label{Data}
Almost all medical prescriptions in French have identical structures. Based on our database and the knowledge of our physicians, we visualized and concluded that 90\% of the prescription could be categorized into three types of formats. The rest have either a unique structure or are handwritten. Hence, this insight helps us choose the right way to achieve the desired goal, which we will develop in the following sections. Figure~\ref{scanned_documents} shows the variety of prescriptions that are the subject of focus.

We had more than 2000 medical documents classified between blood work, medical reports, and prescriptions in our internal databases. 500 of them are medical prescriptions. Upon examination, we found out that only 70 users accepted that their documents could be used in our research work, although they have already gone through our anonymization algorithms. 

Therefore, we launched a campaign to collect data from our work colleagues and our families (47 persons), and we got over 100 prescriptions from 10 different cities and different clinics.
Briefly, for the training dataset, we obtain 170 prescriptions where each of them has at least five drug names followed by 1-3 sentences describing how it should be taken (posology). Practically all of them have a well-defined structure with some changes in the header and layout of the document frame. Again, we can find prescriptions that have a unique format.
%Regarding the testing dataset, we gathered data from another 20 people from our colleague at work. This data was constituted of 33 prescriptions. They contain in total 75 drugs names and 61 posology sentences. Some of the drugs do not have an associated posology, where other have multiple ones.

%based on this document and the expertise of the medicine that they are collaborating with us in this project, we have created more fake document and we have scanned them with different smartphone (to imitate the real life condition) to test our method.

\subsection{Annotation tools}\label{Annotated}
 
For the project development, it was decided to apply a textual classification model to the prescription before extracting the drug-related information. To proceed with this plan, the first step consists of using annotation tools to create the labeling data. This step was carried out by the Prodigy web application developed by the same Spacy team. It is a paid tool. It offers a streaming display of all the document sentences one by one. The data scientist's role is to choose one of the predefined categories for each sentence as a label. Ultimately, this results in a database of all of the sentence-level annotated texts. Prodigy proposes several training functions and a continuous active learning system, but we did not use this system in our project.

%We have just using the annotation tools for the detection of the entity in the posolgy. weh have used a empty spacy NER model to train our NER model. we have used our synthetic data that have been generated via our physician. despite the efficient of this model, but the result was not good enough to just st oping improving the detected entity. Hence, we have created manually more than 100 pattern for each category based on rule have been created by physician. we have included all kind of abbreviation and Shortcuts can be used by the docteur to write the posology. even we have taken into account the grammatical fault that can often be  written and that have the same meaning of the original work without having of misunderstading of the sentence. "comprimeees" is the same as "comprime". any word in the posology has a meaning, we have the same verbes, same adjective, some special caractere. We have given time for this text, and we have succeed to cover the majority of them. 
%add a photo for the pattern that we have created for the drug exemple. we have used the rule-based approach of the Spacy libray to create ou pattern. the advantage of this approach that  you can for each token can set multiple attributes like text value, part-of-speech tag or boolean flags in a easy way with less complexity.

\subsection{Drug names database}\label{Drug}

In this work, we are not just looking to extract the name of the drug from the prescription. One of the main objectives is to link the detected drug to an official reference database where a drug can have a unique ID and a description of its medical information. In addition, we are looking for an international solution as we are working on an international project. Thus, we relied upon two drug databases.  The first one is the Vidal Drug Information Systems database \cite{vidal}. This database contains drug information in several languages, and it can be used for many other functionalities that will be useful for the project in future work. The second database we used is the French Government drug database  \cite{medicaments}. 
%This database has been used to create the model of text classification.

\subsection{Methods}

\begin{figure}[b]
\centerline{\includegraphics[scale=0.32]{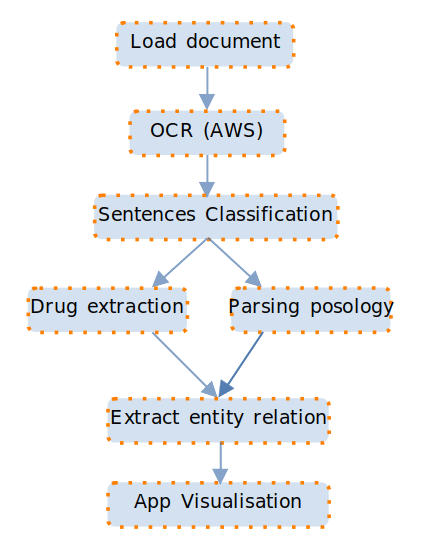}}
\caption{Flowchart of our appraoch.}
\label{algorithms}
\end{figure}

Our objective is to develop a drug-related information extraction system and deploy it in an actual web/mobile application. In this section, we will describe the adopted solution for each task in this project. Figure~\ref{algorithms} shows the sequence of operations that have been applied to obtain the final output.

Since this solution is intended to be used in a real-time application, it is crucial to develop a design with a precise and trusted outcome. This is one of our main contribution. To our knowledge, no one has yet developed such an application. The existing state-of-art is academic, and it is implemented on data from hospitals or clinics, unlike ours, which is composed of scanned documents by the patient using different smartphone cameras. That represents a higher challenge due to the document quality and the versatility that does not exist in other work. 

As shown in Figure~\ref{algorithms}, firstly, we implement the OCR technique to extract text from documents in pdf or image format, which includes the skew correction. Then we apply the deep learning method from the Spacy library to classify sentences between drug, posology, or useless sentences. Depending on the predicted class, if it is a drug sentence, we apply a particular matcher to find out the drug's name and attach it to a unique ID in the Vidal databases. Otherwise, a Spacy rule-based matcher is applied to extract the drug-related information (dosage, frequency, duration, comments) if it is a posology sentence. These matchers are highly customized to fit only French prescriptions that have specifics formats. We follow this step with a geometric relationship approach to assign each posology to its corresponding drug. At the end, we display the result in a structured user interface for a mobile/web application. In the following sections, a detailed description will be presented for each of the proposed tasks.

%In this section we describe our approach for drug entity recognition, drug related-information  recognition, patterns construction and relation extraction before presenting our evaluation method. \newline

\subsection{Optical character recognition}

In the existing research work, most projects operate either on pure medical text or PDF documents. Nobody needed to examine the quality of the documents or process the noise in data to extract the texts. That implies they had no limitation regarding the document's quality or the method used to extract the text. Despite the limited number of documents in our database, there is a diversity in the quality of photos and how the posology is drafted for each drug. Therefore,  the extraction of texts  is the fundamental step to have a clean data to process in the coming phases. For this purpose, we use the service of AWS to carry out this task. We compared it with other open-source methods, but the AWS results were persuasive enough regarding their performance on the text extraction quality with all its details.
This functionality takes as input a document's photo and returns a JSON file containing the text and its geometric-related information in sentence level and word level.

\subsection{Text pre-processing}

Due to the low quality of the scanned document, OCR may return unclean texts. For this reason, a data cleaning step is crucial to make the texts more useful. We have applied the following treatment to the extracted text: 
\begin{itemize}
  \item Remove the accents.
  \item Transfer words to lowercase format.
  \item Unify the format of the number. Remove space and unnecessary punctuation between them if they exist. 
  \item Remove the stop words.
  \item Remove sentences with one word of less than two characters (we got many of them).
\end{itemize}

\subsection{Sentence classification}

Before we started extracting the desired entities from the text, it was more efficient to categorize the text sentences and exclusively use those containing the drug and its information. This step will increase the efficiency of our NER model since it will only focus on sentences that contain the desired entities. Moreover, with limited databases, learning a model to classify the sentence is more realistic and doable than learning a NER model.

To accomplish this task, we downloaded 15000 drug names from the government public drug database. We also generated 15000 synthetic sentences that describe how a drug should be taken (these sentences are called posologies). Moreover, we have produced over 15000 sentences carrying medical information, patient information, names, and all kinds of information that may be present in a prescription other than the drug and the posology.

We trained our classifier using these 45000 sentences with a bidirectional LSTM architecture. It is used to classify sentences into three categories; drug, posology, and useless sentences. This classifier achieved an accuracy of 95.23\%. The remaining 5\% will be automatically ignored by the drug extraction model due to the rules-based approach.

\begin{figure*}[]
\centerline{\includegraphics[scale=0.5]{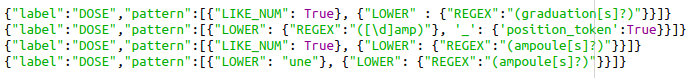}}
\caption{Patterns that have developed using the rule-based matching Spacy library.}
\label{Rules_pattern}
\end{figure*}

\subsection{Drug detection}

In a French prescription, the text is structured in a particular way. A drug is located on a separate line or on the same line with the posology. So, after getting the classified sentence from the previous model, we will know in advance that there are two choices for the sentence classified as "drug sentence"; either this sentence contains only the name of the drug, or it contains both the name of the drug and its related information (posology). In addition, generally, between two drug sentences, there is only one or more posology. We have never seen any information unrelated to posology.

Moreover, one of the main points that generally holds true in French prescriptions is that the names of the drugs are consistently among the top three words in the sentence. That information led us to minimize the search within a window in the sentence. Hence, we have created a drug-matcher that relied on a rule-based approach using the French Government drug databases \cite{medicaments}. Once we have detected the first word of the drug name, it is used to send a query to the Vidal Drug Databases and extract a list of all similar names and features. Then, we apply similarity measurements to determine the closest and longest contiguous matching sub-sequence to the name in the prescription. Indeed, this task is not evident since the name of each drug is composed of up to five words including numbers and units. We can see that clearly in Figure \ref{scanned_documents}. We often obtain the closest match despite the difference in word succession or the spaces between numbers and units.

\subsection{Posology detection}

Similar to the previous section, after the sentence is classified as a posology sentence,  a rule-based matcher will be applied over the sentence to extract the related drug information. 
We used the 170 prescriptions introduced in Section \ref{Data} to produce the related information pattern. We employed the rule-based model from the NLP Spacy library to create the entity matcher. We designed four matchers for the four different features we search for (dosage, frequency, duration, comment). Figure \ref{Rules_pattern}  shows the rules created for the "dose matcher". Our physicians designed more than 100 patterns for each feature. They include abbreviations, miss-writing, and common mistakes.

\begin{table}[b]
\caption{Medication information predictions metrics results.}
\centering
\resizebox{.4\textwidth}{!}{%
\begin{tabular}{llll} 
\hline
Label     & F-measure & Precision  & Recall  \\ 
\hline
Drug name  & 94.33     & 100      & 89.33   \\
Dose       & 93.91     & 100      & 88.52   \\
Duration   & 94.91     & 98.24    & 91.80   \\
Frequency  & 96.60     & 100      & 93.44   \\
Comment    & 91.10     & 100      & 83.60   \\
\hline
\end{tabular}
}
\label{table_result}
\end{table}

\subsection{Drug-posology relation extraction}
The AWS "textract" API output format includes geometric coordinates of the polygons enclosing the words and the sentences. Based on this information, we created an algorithm to assign each entity to its corresponding drugs using the geometric features. We relied on human linguistic intuition in this approach. Each posology is assigned to the closest top drug while respecting a given distance threshold between their polygons. A drug can have a posology composed of several lines. So, for a given text, if the successive sentences respect the given distance threshold, they are considered in the same section and associated to the same drug. Otherwise, when the posology is aligned horizontally with a drug, it will automatically be assigned to it.

\section{Evaluation and Results}\label{EvaluationResult}
In order to evaluate our method, we used the standard metrics for this task. We measure the recall, the precision, and the F1-score on our test dataset for each feature by itself.  An entity is considered a true-positive when it was annotated with the correct label, a false-positive when a token is falsely annotated with respect to each feature. A  false-negative is considered when it was not annotated at all, or it was annotated with the incorrect label\cite{jouffroy2021hybrid}.

Regarding the testing data, we gathered documents from 20 colleagues at work. These 20 colleagues are located in different cities in France and work remotely with the ReLyfe group. The majority are in Paris and Reims.  The data constituted of 33 prescriptions. This data is composed of 1096 sentences with 4572 words. It contains in total 75 drug names and 61 posology sentences. Some of the drugs do not have an associated posology, where others have multiple ones. Despite the limited number of documents for testing, these documents are characterized by their diversity. Each one has a different drug and a different way of representing how to take it (posology). Table \ref{table_result} summarises the results of each different entity matcher. We can see that the model results are higher than 90\%, which is not unexpected due to the high number of patterns that we have created to cover the maximum number of cases and sentences. In addition, as mentioned in the previous section, more than 80\% of the prescription documents have the same structure and format, which helped us get these good results. We can notice that the precision for all models is approximately 100\%. That may be due to the accuracy of the sentence classifier model that is applied beforehand. This model eliminates the confusion presented when we apply our NER models to sentences that do not fit into the posology category.

\begin{figure}[b]
\centerline{\includegraphics[scale=0.2]{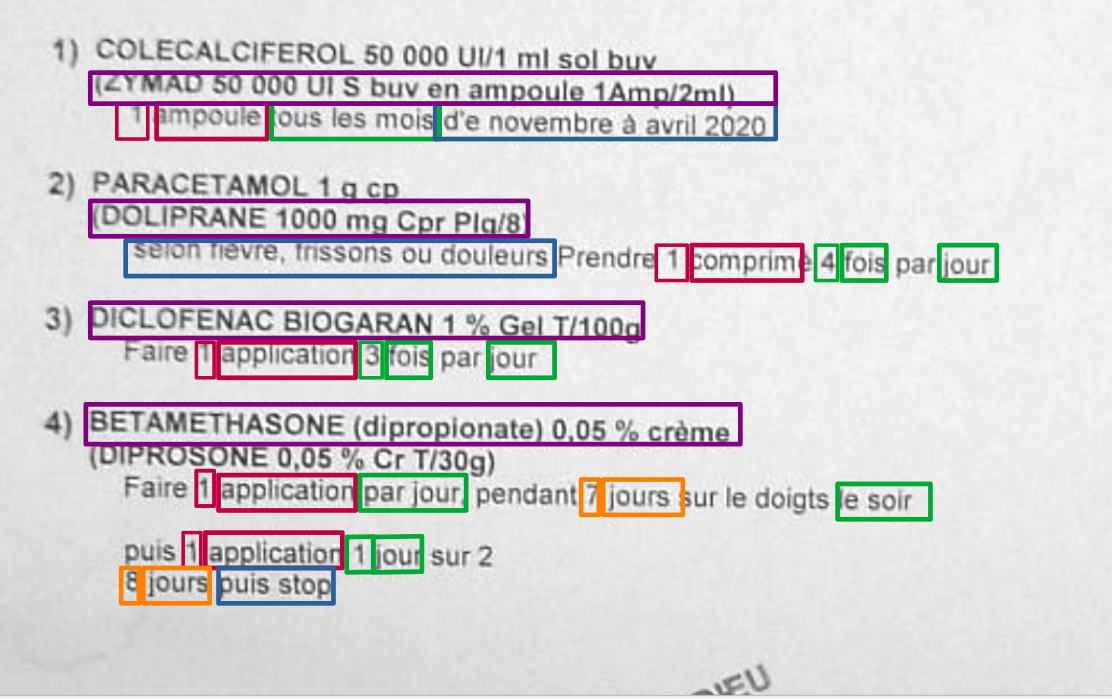}}
\caption{The extracted entities from our approach. Despite the low quality of this photo, we were able to extract the text and apply the relation entity extraction models.}
\label{edited_photo}
\end{figure}

Figure~\ref{edited_photo} shows the results of getting the drugs and their medical-related information from a prescription. This document has seven drug names and four posologies. After getting all the desired entities, they will be the input of another algorithm to differentiate between numbers and units, as shown in the photo.  Moreover, sometimes physicians provide equivalent to each drug if it does not exist, so our algorithm works to choose one of them to make it the reference. Figure~\ref{display_data} displays the extracted entities in our web/mobile application where the user can upload a French prescription and get the structured data.

%\section{Discussion}\label{challenge}
%the precision of the OCR system \newline
%the long name of the drug, where sometime is missed some word \newline
%sections \ref{}--\ref{} below for more information on 
%Fig~\ref{edited_photo}

\begin{figure}[htb]
\centerline{\includegraphics[scale=0.35]{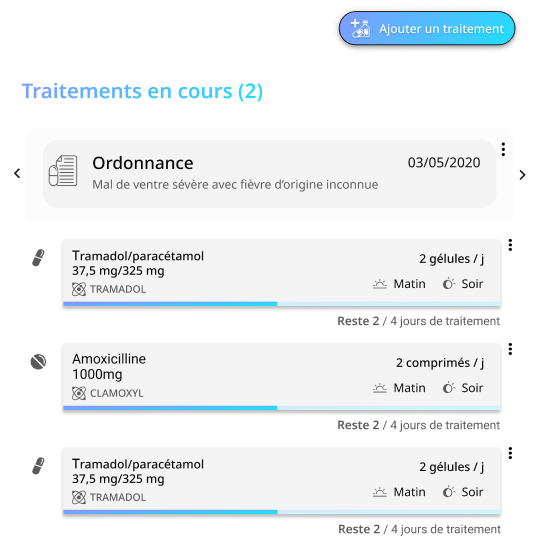}}
\caption{The display of our results in our mobile application after getting the drug and its related information.}
\label{display_data}
\end{figure}

\section{Conclusion}\label{conclusion}

This paper presents a functionality created to be applied in a real application to extract drug-related information. This work is one of the few that contributes to the French medical documents. Our approach is a series of methods concatenated together to achieve a high-performance system capable of coping with the constraints of real applications. We have applied a deep learning technique to classify the prescription sentences into three categories: drug, posology, and useless sentences. Then, we applied the corresponding rule-based approach to extract the targeted features for each of these categories. A unique international ID is associated with the detected drug name via the Vidal databases. Lastly, an algorithm based on human intuition and the sentence's geometric position was designed to build a relation extractor system to associate the detected entities to their corresponding drug. Theoretical and practical tests have proved the outperformance of our approach.

\bibliographystyle{IEEEtran}
\bibliography{biblio}

\end{document}